\documentclass[letterpaper]{article} 
\usepackage{aaai25}  
\usepackage{times}  
\usepackage{helvet}  
\usepackage{courier}  
\usepackage[hyphens]{url}  
\usepackage{graphicx} 
\usepackage{natbib}  
\usepackage{caption} 
\usepackage{algorithm}
\usepackage{algorithmic}
\usepackage{amsmath}
\usepackage{amssymb}
\usepackage{multirow}
\usepackage{capt-of,etoolbox}

\usepackage[dvipsnames,table]{xcolor}
\usepackage{newfloat}
\usepackage{booktabs}
\usepackage{fontawesome}
\usepackage{listings}
\DeclareCaptionStyle{ruled}{labelfont=normalfont,labelsep=colon,strut=off} 
\floatstyle{ruled}
\newfloat{listing}{tb}{lst}{}
\floatname{listing}{Listing}
\title{Human-VDM: Learning Single-Image 3D Human Gaussian Splatting \\ from Video Diffusion Models}
\author{
    Zhibin Liu\textsuperscript{\rm 1}, Haoye Dong\textsuperscript{\rm 2}, Aviral Chharia\textsuperscript{\rm 2}, Hefeng Wu\textsuperscript{\rm 1}\\
}
\affiliations{
    \textsuperscript{\rm 1}Sun Yat-Sen University \ \ \textsuperscript{\rm 2}Carnegie Mellon University\\
    liuzhb26@mail2.sysu.edu.cn, \{haoyed, achharia\}@andrew.cmu.edu, wuhefeng@mail.sysu.edu.cn\\
    {\color{purple}{\texttt{https://Human-VDM.github.io/Human-VDM}}}
}

\definecolor{bestcolor}{rgb}{1, 0.7, 0.5}
\definecolor{secondbestcolor}{rgb}{1, 0.9, 0.7}
\newcommand{\bone}{\cellcolor{bestcolor}}
\newcommand{\btwo}{\cellcolor{secondbestcolor}}
\usepackage{bibentry}

\begin{document}

\twocolumn[{
\renewcommand\twocolumn[1][]{#1}
\maketitle
\vspace{-1em}
\begin{center}
    \captionsetup{type=figure}
    \includegraphics[width=1.0\textwidth]{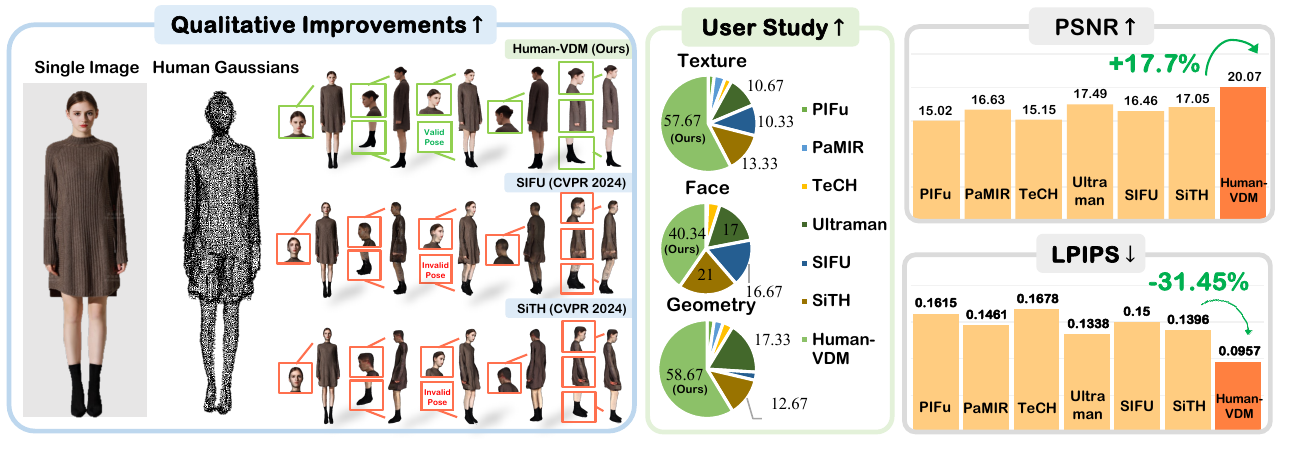}
    \vspace{-0.75cm}
    \captionof{figure}{\textbf{Human-VDM for generating 3D humans from a single image.} Given a single RGB human image, Human-VDM aims to generate high-fidelity 3D human. Human-VDM preserves face identity, delivers realistic texture, ensures accurate geometry, and maintains a valid pose of the generated 3D human, surpassing the current state-of-the-art models.}
    \vspace{1em}
    \label{Teasor}
\end{center}
}]

\begin{abstract}
Generating lifelike 3D humans from a single RGB image remains a challenging task in computer vision, as it requires accurate modeling of geometry, high-quality texture, and plausible unseen parts. Existing methods typically use multi-view diffusion models for 3D generation, but they often face inconsistent view issues, which hinder high-quality 3D human generation. To address this, we propose \textbf{Human-VDM}, a novel method for generating 3D \textbf{human} from a single RGB image using \textbf{V}ideo \textbf{D}iffusion \textbf{M}odels. Human-VDM provides temporally consistent views for 3D human generation using Gaussian Splatting. It consists of three modules: a view-consistent human video diffusion module, a video augmentation module, and a Gaussian Splatting module. First, a single image is fed into a human video diffusion module to generate a coherent human video. Next, the video augmentation module applies super-resolution and video interpolation to enhance the textures and geometric smoothness of the generated video. Finally, the 3D Human Gaussian Splatting module learns lifelike humans under the guidance of these high-resolution and view-consistent images. Experiments demonstrate that Human-VDM achieves high-quality 3D human from a single image, outperforming state-of-the-art methods in both generation quality and quantity.
\end{abstract}

\section{Introduction}
Generating 3D humans from a single RGB image has gained significant attention in recent years due to its versatile applications in filmmaking, video games, human-robotic interaction, etc. However, existing approaches for 3D human generation largely rely on multi-view diffusion models, which often suffer from inconsistent views and lead to artifacts. To address this problem, we propose a 3D Human Gaussian Splatting framework that allows users to generate 3D humans from a single 2D image input while ensuring accurate geometry and realistic appearance. However, generating 3D humans using only a single RGB image presents a significant challenge due to its inherent ambiguity, which necessitates inferring unseen geometry and appearance that are not directly captured in a 2D image.

Current approaches address this challenge by incorporating parametric human shape models, such as SCAPE~\cite{anguelov2005scape} and SMPL~\cite{loper2023smpl}. However, these methods exclusively focus on reconstructing the human shape, neglecting the appearance details crucial for a fully realistic 3D representation. Earlier works, like PIFu~\cite{saito2019pifu}, attempted to address this gap with a data-driven approach. They used CycleGAN~\cite{zhu2017unpaired} and residual blocks~\cite{johnson2016perceptual} trained on image-3D pairs. However, such methods often struggle with novel appearances or poses mainly due to the lack of sufficient 3D training information. Subsequent methods, such as ECON~\cite{xiu2023econ} and 2K2K~\cite{han2023high}, enhanced performance by incorporating depth or normal estimation into the generation process. SIFU~\cite{zhang2024sifu} proposed a 3D human generation method using a side-view based Transformer with 3D aware Refinement. Despite the improvements, these methods often lack detail or result in inaccurate geometry, particularly with high-resolution input images.

Recently, SiTH~\cite{ho2024sith} integrated a generative diffusion model into the 3D human generation pipeline to produce realistic textures and geometries, especially in unobserved regions. Ultraman~\cite{chen2024ultraman} introduced a multi-view image generation model that helped in providing essential appearance priors aiding the generation process. Although diffusion models~\cite{rombach2022high}, trained on extensive image datasets, have demonstrated potential for creating 3D humans, multi-view diffusion often struggles with generating view-consistent images and tends to introduce artifacts in the generated 3D humans.

This paper proposes Human-VDM, a novel Gaussian Splatting framework for generating 3D humans from a single image using video diffusion models. Human-VDM is comprised of three distinct modules: a view-consistent human video diffusion module, a video augmentation module, and a 3D human Gaussian Splatting module. Human-VDM first generates a `view-consistent' human video, then enhances the quality of the frames through super-resolution and video frame interpolation, and finally employs 3D Gaussian Splatting (3DGS)~\cite{kerbl20233d} to effectively generate the 3D human model. 

Initially, we fine-tune SV3D~\cite{voleti2024sv3d}, a latent video diffusion model specifically designed for generating object videos, to enable it to generate view-consistent human videos. However, a direct application of video diffusion models~\cite{voleti2024sv3d} to the 3D human generation can result in geometric artifacts and blurry textures. Additionally, the generated video consists of only 21 frames at a low resolution of $576 \times 576$, which is insufficient for high-quality 3D human generation. To provide more view-consistent frames and realistic texture for 3D human generation, we carefully designed a video augmentation module that includes super-resolution and frame interpolation components. The generated human video is enhanced through this module by undergoing super-resolution and frame interpolation, which results in smooth, high-quality frames at a resolution of $1080 \times 1080$. Lastly, we introduce a 3D human Gaussian splatting module to generate realistic 3D human models. For this, we utilize SMPL~\cite{loper2023smpl} along with an optimizable feature tensor training strategy to optimize the parameters of the 3D Gaussians, thereby generating a high-quality 3D human from a single image. Figure~\ref{Teasor} and~\ref{performance} demonstrate that Human-VDM achieves state-of-the-art (SOTA) performance and generates realistic 3D humans from a single-view RGB image input. Our contributions can be summarized as follows:
\begin{itemize}
\item We propose a novel single-view 3D human generation framework that leverages the human video diffusion model to produce view-consistent human frames.
\item We carefully designed a video augmentation model that consists of super-resolution and video frame interpolation to enhance the quality of the generated video.
\item We introduce an effective Gaussian Splatting framework for 3D human reconstruction with offset prediction.
\item Extensive experiments demonstrate that the proposed Human-VDM can generate realistic 3D humans from single-view images, outperforming state-of-the-art methods in both quality and effectiveness.
\end{itemize}

\begin{figure*}[t]
\centering
\includegraphics[width=1\textwidth]{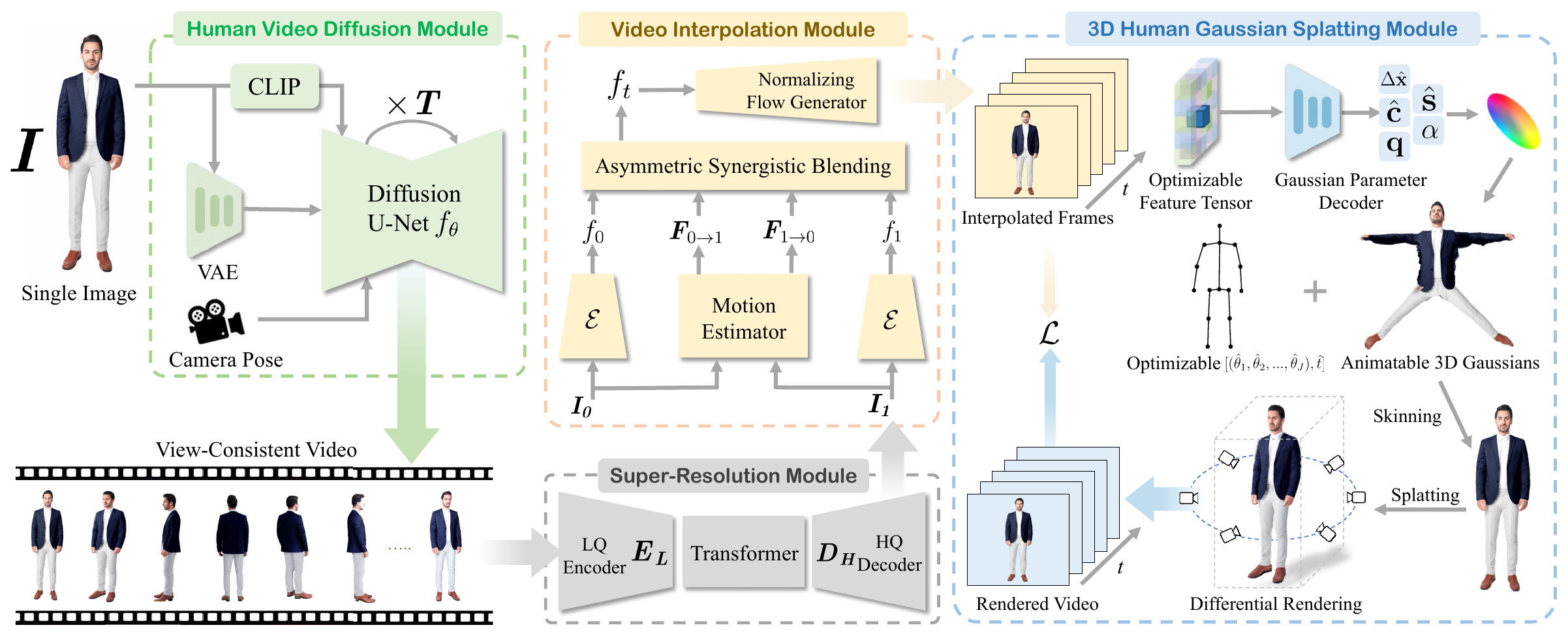}
\caption{\textbf{Human-VDM model architecture.} An image $I$ is first input to a view-consistent human video diffusion module to generate a coherent human video. Next, the video augmentation module applies super-resolution and frame interpolation to enhance texture and generate high-quality interpolated frames. Finally, 3D Human Gaussian splatting learns lifelike 3D humans.}
\label{pipeline}
\end{figure*}

\section{Related Works} 

\textbf{3D Human Generation.} PIFu~\cite{saito2019pifu} was among the first methods to introduce pixel-aligned features and neural fields~\cite{xie2022neural} for reconstructing human figures from images by fitting parametric human shape models such as SMPL~\cite{loper2023smpl} and SCAPE~\cite{anguelov2005scape}. PIFuHD~\cite{saito2020pifuhd} further enhanced this framework with high-resolution normal guidance. Subsequent methods improved upon this initial approach by integrating additional human body priors. For instance, PaMIR~\cite{zheng2021pamir} and ICON~\cite{xiu2022icon} utilized skinned body models to guide the reconstruction process, while ARCH~\cite{huang2020arch}, ARCH++~\cite{he2021arch++}, and CAR~\cite{liao2023high} extended this approach by mapping global coordinates into canonical coordinates, enabling reposing. PHOHRUM~\cite{alldieck2022photorealistic} and S3F~\cite{corona2023structured} introduced techniques to disentangle shading and albedo, facilitating relighting. Concurrently, another set of methods replaced neural representations with traditional Poisson surface reconstruction~\cite{kazhdan2013screened}. Despite these advancements, such approaches have been primarily tailored to human bodies and often struggle with the complex topologies of loose clothing. To address this limitation, ECON~\cite{xiu2023econ} and 2K2K~\cite{han2023high} integrated depth or normal estimation to enhance the reconstruction process. More recently, Ultraman~\cite{chen2024ultraman} introduced a model to map texture  thereby optimizing the texture details thus helping to maintain the color consistency during the final reconstruction. SIFU~\cite{zhang2024sifu} also proposed a novel approach that combined the 3D Consistent Texture Refinement pipeline with a side-view Decoupling Transformer.
\vspace{0.5em}

\noindent \textbf{3D Human Generation with Diffusion models.} Diffusion models~\cite{ramesh2022hierarchical} trained on large image datasets have exhibited remarkable capabilities in generating 3D objects from text prompts. Earlier works, such as Fantasia3d~\cite{chen2023fantasia3d} and Magic3d~\cite{lin2023magic3d}, predominantly followed an optimization-based workflow where 3D representations, such as NeRF~\cite{mildenhall2021nerf}, were updated through neural rendering~\cite{tewari2022advances}. Although a few studies, such as TeCH~\cite{huang2024tech}, adapted this workflow for 3D human reconstruction, they struggled to achieve accurate appearance and geometric representations of the human body due to the inherent ambiguities in text prompt condition. Recently, SiTH~\cite{ho2024sith} integrated a generative diffusion model to produce full-body texture and geometry, including unobserved regions, within the reconstruction workflow. However, these methods still face challenges in capturing detailed clothing. In this paper, we leverage a video diffusion model (VDM) to generate an orbital video for 3D human reconstruction.

\section{Human-VDM}

Given a single RGB image $I$ of a person, Human-VDM aims to generate its 3D human model (see Figure~\ref{pipeline}). Human-VDM comprises several key modules: (i) the Human Video Diffusion module, (ii) the Video augmentation module, which includes the super-resolution and frame interpolation sub-modules, and (iii) the Human Gaussian Splatting module. First, the Human Video Diffusion module generates view-consistent videos of the input image. This video is then processed by the Video Augmentation module, where super-resolution enhances the resolution to $1080\times1080$, while video frame interpolation (VFI) smoothens the video frames. Finally, the augmented video is fed into the Human Gaussian Splatting module to generate a high-fidelity 3D human model.

\subsection{Human Video Diffusion Module}

To generate the video $\hat{V}$, we input the front image of a human, denoted as $I$, into a latent video diffusion model which we fine-tuned for high-quality human video generation. We specifically use SV3D~\cite{voleti2024sv3d}, a latent video diffusion model designed for generating videos from a single image, capable of producing consistent multi-view images. However, since SV3D was originally designed for reconstructing general objects, its generated video quality for human body images is not satisfactory. Therefore, to enhance its capability for human video generation, we fine-tuned SV3D on Thuman~2.0~\cite{tao2021function4d} dataset which includes a variety of high-quality human body scans. SV3D produces a raw orbital video, $\hat{V} = [\hat{f}_1, \hat{f}_2, \hat{f}_3, \ldots, \hat{f}_{21}]$, with a resolution of $576 \times 576$, illustrating the human from different viewpoints. The videos generated by the fine-tuned SV3D exhibit superior shape, appearance, and detailed rendering of areas not directly captured in a 2D image. We represent this generation process as follows:
\begin{equation} \label{f0}
\begin{split}
    \hat{V} = \text{SV3D}(I),
\end{split}
\end{equation}
where `SV3D' denotes the generative process of the fine-tuned SV3D model.

\begin{figure*}[!h]
\centering
\includegraphics[width=1\textwidth]{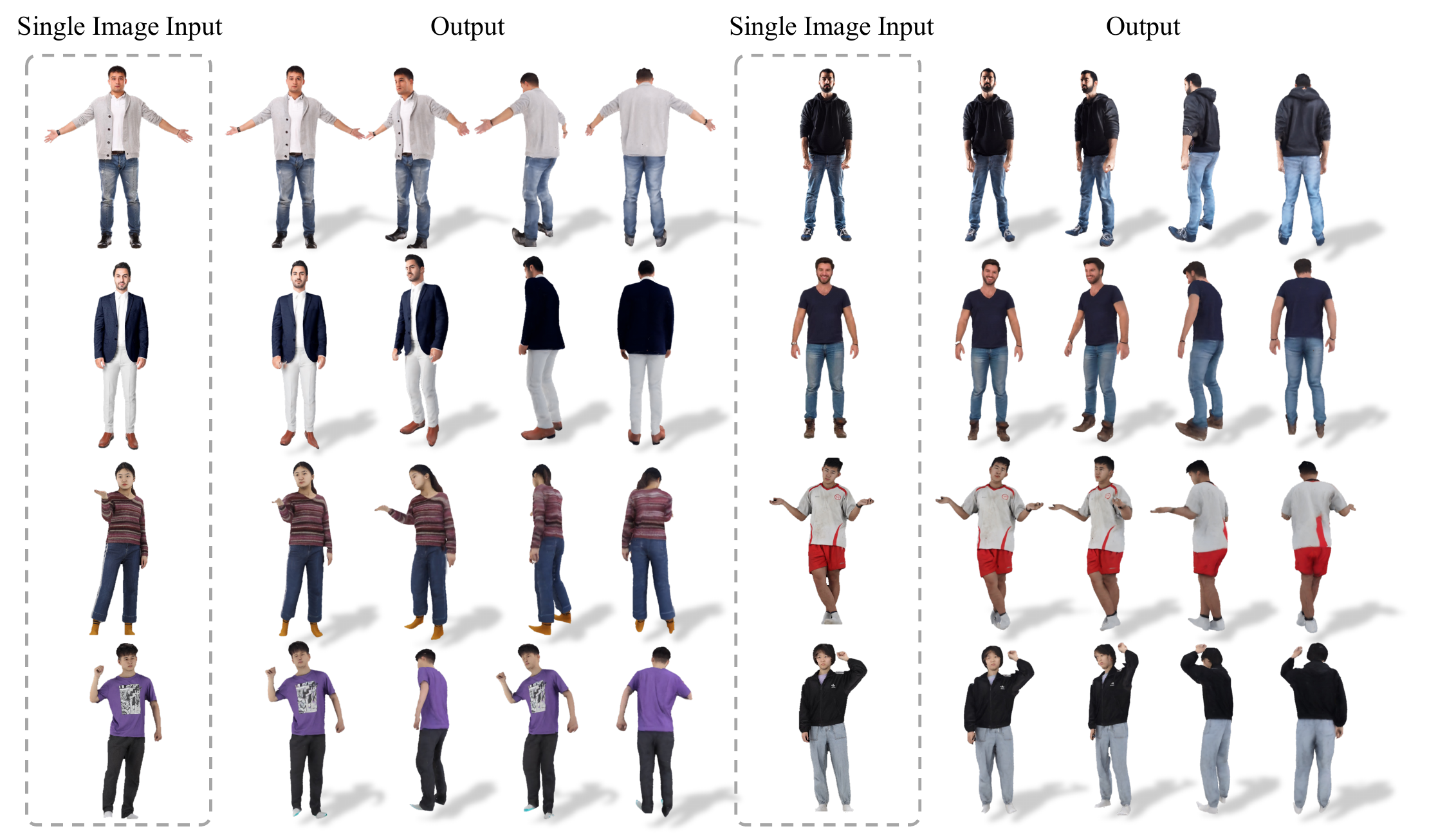}
\caption{\textbf{Qualitative Results.} Novel view results from Human-VDM with various poses, genders, diverse clothing, and different hairstyles demonstrate the robustness of the proposed Human-VDM model. It consistently achieves high photo-realistic quality and precise geometric accuracy. \faSearch~\textbf{zoom in} for details.}
\label{performance}
\end{figure*}

\subsection{Video Augmentation Module}

The 21-frame human video $\hat{V}$, with a resolution of $576 \times 576$, has limited expressive capacity for detailed 3D human reconstruction. To address this, we introduce the Video Augmentation Module, which includes super-resolution and frame interpolation. Super-resolution helps in improving the quality of textures while video frame interpolation improves the geometric smoothness of the 3D human and the quality of the previously invisible areas.
\vspace{0.5em}

\noindent \textbf{Video Super-resolution sub-module.} For image super-resolution on each frame of $\hat{V}$, we employ CodeFormer~\cite{zhou2022towards}, a transformer-based model designed primarily for enhancing facial image resolution. CodeFormer performs Low Quality (LQ) to High Quality (HQ) mapping by first learning a discrete codebook and an HQ decoder $D_H$ through self-reconstruction learning. During Codebook Lookup, a transformer and an LQ encoder $E_L$ are additionally introduced to accurately model the cookbook code combination. For facial images, increasing the resolution of each frame of $\hat{V}$ by $4\times$ and then resizing it to $1080\times1080$ yields clear and realistic images that significantly benefit 3D reconstruction. Similarly, we increase the resolution of each frame in the raw orbital video $\hat{V}$ by $4\times$ and resize it to $1080\times1080$, resulting in a high-resolution video $V^{'}=[f^{'}_1,f^{'}_2,...,f^{'}_{21}]$ with improved texture quality. This process is formulated as follows:
\begin{equation} \label{f1}
    f^{'}_i = \text{Resize}(\text{SuperResolution} (\hat{f}_i)), \ \ 1\leq i\leq 21,
\end{equation}
where `SuperResolution' denotes the operation of CodeFormer, while `Resize' denotes the operation of resizing the image to $1080\times1080$.
\vspace{0.5em}

\noindent \textbf{Video Frame Interpolation (VFI) sub-module.} To enhance video consistency and interpolate frames, we employ PerVFI~\cite{wu2024perception}. VFI provides additional visual information from diverse angles, improving the geometric smoothness of the 3D human and the quality of the invisible areas. PerVFI performs perception-oriented VFI and inputs two reference frame images $I_0$ and $I_1$ to reconstruct intermediate frames. First, bidirectional optical flows, i.e., $F_{0\rightarrow1}$ and $F_{1\rightarrow0}$ are estimated using a motion estimator. Additionally, two encoders capture multi-scale features. These features are then blended using asymmetric synergistic blending to obtain intermediate features $f_t$. These features are finally decoded to obtain the intermediate frame using a conditional flow generator, which samples from a normal distribution. We input the 21-frame high-resolution video frames $V^{'}$ into PerVFI, resulting in an 81-frame high-resolution augmented video $V=[f_1,f_2, ..., f_{81}]$. This is formulated as follows:
\begin{equation} \label{f2}
\begin{split}
    f = \text{VFI}(f^{'}_j), \ \ 1\leq j\leq 81,
\end{split}
\end{equation}
where `VFI' denotes the frame interpolation operation.

\begin{figure*}[t]
\centering
\includegraphics[width=1\textwidth]{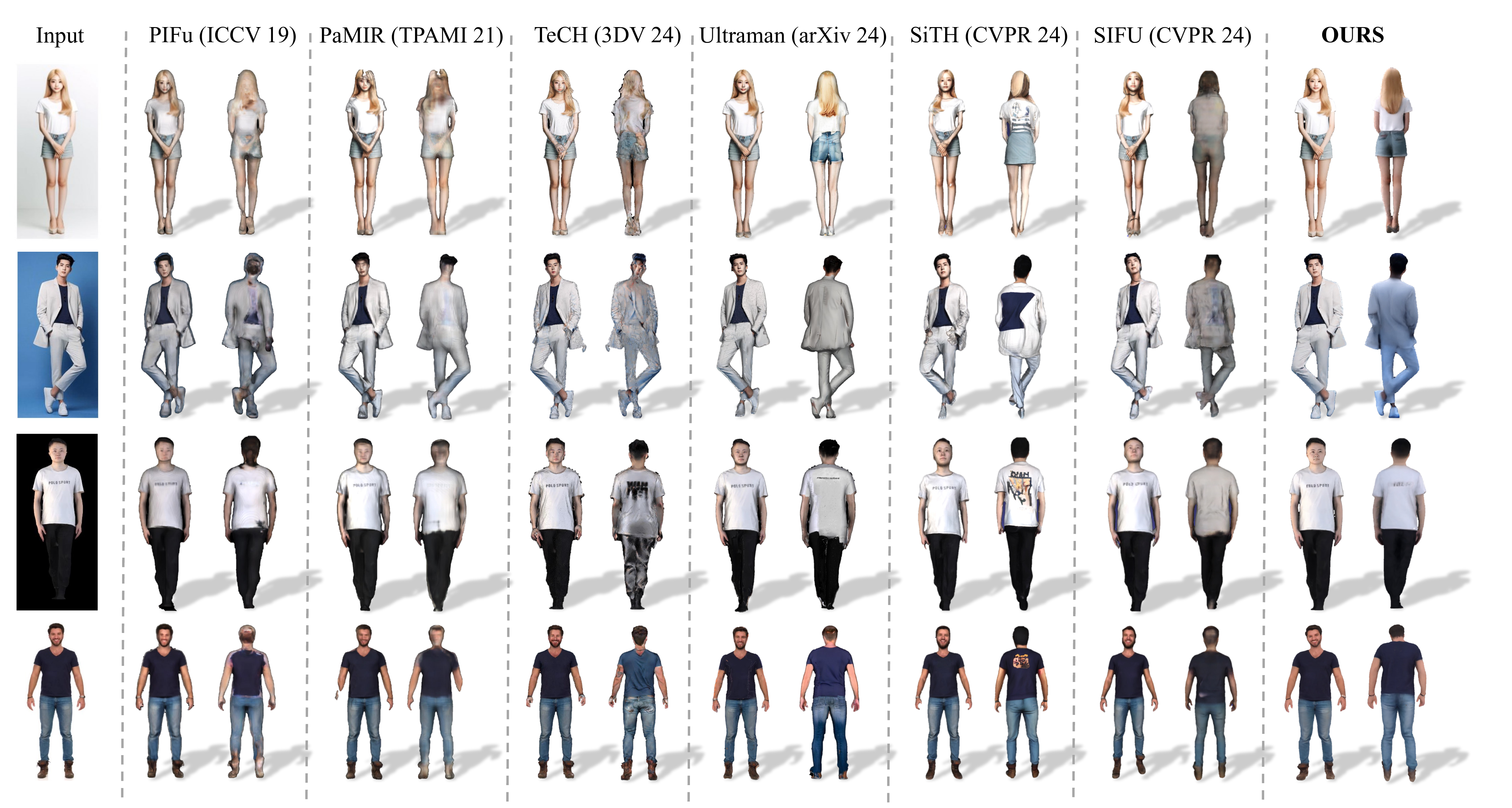}
\caption{\textbf{Qualitative Comparison.} Human-VDM compared to other SOTA models including PIFu~\cite{saito2019pifu}, PaMIR~\cite{zheng2021pamir}, TeCH~\cite{huang2024tech}, Ultraman~\cite{chen2024ultraman}, SiTH~\cite{ho2024sith}, and SIFU~\cite{zhang2024sifu}. The results demonstrate that Human-VDM achieves superior 3D human generation quality. Note that recent SOTAs fail to predict the unseen back view as shown above. \faSearch~\textbf{zoom in} for details.}
\label{comapre}
\end{figure*}

\subsection{3D Human Gaussian Splatting Module}
We leverage 3D Gaussian Splatting~\cite{kerbl20233d} to model the 3D human from the augmented human video $V$. 3D Gaussian Splatting employs point-based representation, which facilitates high-quality real-time rendering by modeling the 3D object as a collection of parameterized static 3D Gaussians. Each Gaussian is characterized by a color $c \in \mathbb{R}^3$, a 3D center position $x \in \mathbb{R}^3$, opacity $\alpha \in \mathbb{R}$, a 3D scaling factor $s \in \mathbb{R}^3$, and a 3D rotation $q \in \mathbb{R}^4$.

In this module, we incorporate an appearance network in conjunction with an optimizable feature tensor to enhance the representation of 3D Gaussian models refined from video data~\cite{hu2024gaussianavatar}. For each $i^{th}$ frame $f_i$ in the augmented video $V$, we first extract the SMPL model of the human body. We then sample points on the surface of this model and map their positions onto a UV position map, denoted by $m$. We introduce an optimizable feature tensor to capture the appearance of the reconstructed human. The parameters for each Gaussian are predicted by a Gaussian parameter decoder using the optimizable feature concatenated with $m$ as input. These predictions form the 3D Gaussians in the canonical space. Using Linear Blend Skinning (LBS), these canonical 3D Gaussians can be reposed into motion space for rendering. This is formulated as follows:
\begin{equation} \label{f3}
\begin{split}
    m &= M(\tilde{\theta},\beta) \\
    P &= Decode(cat(t,m)),\\
    f_i^r &= \text{Splatting}(\text{LBS}(D,J(\beta),\hat{\theta}_i),P),
\end{split}
\end{equation}
where $\tilde{\theta}$ is the pose parameters of the SMPL model in canonical space and $\beta$ is the average shape parameters calculated from $V$, respectively. $M$ is the operation of mapping the positions of the sampled points on the surface of the SMPL model onto a UV map; $t$ denotes the optimizable feature tensor, $Decode$ means the process of decoding the aligned feature tensors to predict the parameters of Gaussians $P$. $D = T(\beta) + dT$ denotes the locations of 3D Gaussians in canonical space, formed by adding corrective point displacements dT on the template mesh surface $T(\beta)$, $J(\beta)$ produces 3D joint locations, $\hat{\theta}_i$ represents the refined pose parameter optimized from $\theta_i$, which denotes the pose parameters obtained from $f_i$, `LBS' is the operation of Linear Blend Skinning; `Splatting' denotes the render process, resulting in a rendered image $f_i^r$. 
\vspace{0.5em}

\begin{table*}[!h]
\renewcommand{\arraystretch}{1.05}
\setlength{\tabcolsep}{2pt}
\centering
\caption{\textbf{User study and Quantitative Comparisons.} Human-VDM compared to recent single-image based 3D human generation SOTAs. Top two results are colored as \colorbox{bestcolor}{first}~\colorbox{secondbestcolor}{second}.}
\resizebox{1\textwidth}{!}{
\begin{tabular}{lc|cccc|cccc}
\hline
\multirow{3}{*}{Method} & \multirow{3}{*}{Venue} & \multicolumn{4}{c|}{User Study} & \multicolumn{4}{c}{Quantitative Evaluation}\\ 
\cline{3-10} 

& & \multirow{2}{*}{Geometry~(\%)} & \multirow{2}{*}{Texture~(\%)} & \multirow{2}{*}{Face~(\%)} & Which is & \multirow{2}{*}{CLIP Sim. $\uparrow$} & \multirow{2}{*}{SSIM $\uparrow$} & \multirow{2}{*}{LPIPS $\downarrow$} & \multirow{2}{*}{PSNR $\uparrow$} \\ 

& & & & & best~(\%) & & & & \\ 

\hline
PIFu~\cite{saito2019pifu} & ICCV~2019 & 2.33 & 2.00 & 0.33 & 1.67 & 0.8501 & 0.8884 & 0.1615 & 15.0248 \\
PaMIR~\cite{zheng2021pamir} & TPAMI~2021 & 3.00 & 3.67 & 0.33 & 2.33 & 0.8861 & 0.8924 & 0.1461 & 16.6267 \\
TeCH~\cite{huang2024tech} & 3DV~2024 & 3.33 & 2.33 & 4.33 & 4.00 & 0.8875 & 0.8709 & 0.1678 & 15.1464 \\
Ultraman~\cite{chen2024ultraman} & arXiv~2024 & \btwo{17.33} & 10.67 & 17.00 & 11.00 & \btwo{0.9131} & 0.8958 &\btwo{0.1338} & \btwo{17.4877} \\
SIFU~\cite{zhang2024sifu} & CVPR~2024 & 2.67 & 10.33 & 16.67 & \btwo{15.67} & 0.8663 & 0.7931 & 0.1500 & 16.4600 \\
SiTH~\cite{ho2024sith} & CVPR~2024 & 12.67 & \btwo{13.33} & \btwo{21.00} & 11.67 & 0.8978 & \btwo{0.8963} & 0.1396 & 17.0533 \\
\hline
\multirow{1}{*}{\textbf{Human-VDM}} & \textbf{Ours} & \bone{\textbf{58.67}} & \bone{\textbf{57.67}} & \bone{\textbf{40.34}} & \bone{\textbf{53.66}} & \bone{\textbf{0.9235}} & \bone{\textbf{0.9228}} & \bone{\textbf{0.0957}} & \bone{\textbf{20.068}} \\
\hline
\end{tabular}
}
\label{user_tab}
\end{table*}

\begin{table*}[!h]
\centering
\caption{\textbf{Ablation studies.} Human-VDM's ablation experiments to verify the effect of proposed components. Without is abbreviated as `w/o'.}
\renewcommand{\arraystretch}{1}
\setlength{\tabcolsep}{3pt}
\resizebox{\textwidth}{!}{
\begin{tabular}{c|cccc|cccc} 
\hline
Ablation & CLIP Sim.$\uparrow$ & SSIM~$\uparrow$ & LPIPS~$\downarrow$ & PSNR~$\uparrow$ & \begin{tabular}[c]{@{}c@{}}CLIP Sim. \\ (Front View)\end{tabular}$\uparrow$ & \begin{tabular}[c]{@{}c@{}}SSIM \\ (Front View)\end{tabular}~$\uparrow$ & \begin{tabular}[c]{@{}c@{}}LPIPS \\ (Front View)\end{tabular}~$\downarrow$ & \begin{tabular}[c]{@{}c@{}}PSNR \\ (Front View)\end{tabular}~$\uparrow$ \\ \hline

w/o frame interpolation & \btwo 0.9234 & \bone 0.9216 & 0.0973 & 20.030 & 0.9286 & \btwo 0.9122 & 0.0930 & 19.75 \\
w/o super-resolution & 0.9231 & 0.8981 & \bone 0.0865 & \bone 20.076 & 0.9448 & 0.8857 & \bone 0.0767 & 19.615 \\
w/o fine-tuned SV3D & 0.9146 & 0.9145 & 0.1062 & 18.726 & \btwo 0.9449 & 0.9095 & 0.0933 & \btwo 19.615 \\
Full & \bone 0.9235 & \bone 0.9228 & \btwo 0.0957 & \btwo 20.068 & \bone 0.9607 & \bone 0.9257 & \btwo 0.0846 & \bone 21.184 \\ 
\hline
\end{tabular}
}
\label{ablation_tab}
\end{table*}

\noindent \textbf{Training Objectives.} For formulating the loss function, we take the current frame image $f_i$ as the ground truth and calculate the loss with the rendered image $f_i^{r}$ for optimization. This is formulated as follows:
\begin{equation} \label{f4}
    \begin{split}
        \mathcal{L} &= \lambda_{\text{RGB}}\mathcal{L}_{\text{RGB}} + 
        \lambda_{\text{SSIM}}\mathcal{L}_{\text{SSIM}} + \lambda_{\text{LPIPS}}\mathcal{L}_{\text{LPIPS}}\\
        & +\lambda_{\text{Offset}}\mathcal{L}_{\text{Offset}}
        + \lambda_{\text{Scale}}\mathcal{L}_{\text{Scale}}
        + \lambda_{f}\mathcal{L}_{f},
    \end{split}
\end{equation} 
where $\mathcal{L}_{\text{RGB}}$ is the L1-loss between the ground truth and the rendered frame. $\mathcal{L}_{\text{SSIM}}$ and $\mathcal{L}_{\text{LPIPS}}$ denotes the SSIM and LPIPS losses, respectively. $\mathcal{L}_{\text{Offset}}$, $\mathcal{L}_{\text{Scale}}$ and $\mathcal{L}_{f}$ calculate the L2-norm of predicted offsets and scales, and the feature map, respectively. The weight coefficients $\lambda_{\text{RGB}}$, $\lambda_{\text{SSIM}}$, $\lambda_{\text{LPIPS}}$, $\lambda_{\text{Offset}}$, $\lambda_{\text{Scale}}$ and $\lambda_{f}$, are set to $0.8$, $0.2$, $0.2$, $10$, $1.0$ and $1.0$ respectively.

\section{Experiments and Results}
\label{exp}

\noindent\textbf{Dataset.} Most works use the popular Thuman~2.0 dataset~\cite{tao2021function4d}, which comprises 2,500 high-quality human body scans, each accompanied by a detailed 3D model and texture mapping. The dataset includes a wide range of action poses and provides the SMPL-X~\cite{pavlakos2019expressive} parameters along with corresponding grids.

\noindent\textbf{Evaluation Metrics.} Following previous works on 3D human generation, we use the four major metrics to evaluate the performance of Human-VDM. These include CLIP-Similarity~\cite{radford2021learning}, LPIPS (Learned Perceptual Image Patch Similarity)~\cite{zhang2018unreasonable}, SSIM~\cite{wang2004image} and PSNR. CLIP~\cite{radford2021learning} measures the similarity between two images, providing a more representative evaluation of image feature similarity. LPIPS~\cite{zhang2018unreasonable}, measures differences based on learned perceptual image patch similarity, aligning more closely with human perception. Likewise, SSIM (Structural Similarity Index)~\cite{wang2004image} is used to compare the luminance, contrast, and structure between two images. Lastly, PSNR (Peak Signal-to-Noise Ratio) assesses image quality based on pixel-level error, making it an error-sensitive evaluation metric.

\noindent\textbf{Training details.} To produce high-quality human videos, we fine-tuned SV3D using the Thuman~2.0 dataset~\cite{tao2021function4d} to enhance its 3D human video generation capabilities. We selected 475 samples from Thuman~2.0, excluding those used in subsequent quantitative comparisons. For each sample, 21 images were rendered from various angles following~\cite{xiu2022icon}. All images corresponding to a sample are rendered at the same horizontal position with a constant angular interval of $360/21$ degree to ensure the consistency of rendered multi-view images. The first rendered image of each body was employed as the input, while the remaining images served as ground truth for fine-tuning SV3D. We freeze the image encoder and decoder of the original SV3D~\cite{voleti2024sv3d} model and optimize the U-Net weights~\cite{ronneberger2015u}. The learning rate was set to \texttt{5e-6} and fine-tuned on one NVIDIA A800 GPU with a batch size of 13.

\subsection{Qualitative Comparison} 
Figure~\ref{performance} presents the qualitative 3D human generation results from Human-VDM on a variety of input images that differ in gender, body posture, lighting, color, and clothing styles. The results demonstrate Human-VDM's significant performance with high appearance consistency, texture, and geometry qualities. Next, we compare Human-VDM with recent SOTA works on single-image based 3D human generation (see Figure~\ref{comapre}), including PIFu~\cite{saito2019pifu}, PaMIR~\cite{zheng2021pamir}, TeCH~\cite{huang2024tech}, Ultraman~\cite{chen2024ultraman}, SiTH~\cite{ho2024sith} and SIFU~\cite{zhang2024sifu}. Compared to Human-VDM, PaMIR~\cite{zheng2021pamir} exhibits significant shortcomings in the geometry of the generated 3D human, e.g., the body of the generated human is incomplete for the first image. On the other hand, TeCH~\cite{huang2024tech}, PIFu~\cite{saito2019pifu}, and SiTH~\cite{ho2024sith} reconstruct remarkable geometries but contain apparent artifacts. Likewise, SIFU~\cite{zhang2024sifu} displays misalignment in character motion and suboptimal texture quality on the back of the generated human. While Ultraman~\cite{chen2024ultraman} obtains good geometry but fails to predict the realistic appearance of unseen view. Therefore, the proposed Human-VDM outperforms SOTA models in terms of texture quality and appearance consistency.

\subsection{Quantitative Comparison} 
\label{quan}
Following previous methods~\cite{chen2024ultraman}, we randomly selected 50 samples from Thuman~2.0~\cite{tao2021function4d}. Four views of the ground truth (GT), i.e., front, back, left, and right, were used to compute scores between the reconstructed results and the GT across these views. As reported in Table~\ref{user_tab}, Human-VDM achieves the lowest LPIPS and the highest CLIP score, indicating that the rendered images produced by our method are highly consistent with the input images. Additionally, Human-VDM achieves the highest SSIM and PSNR scores, further demonstrating that the rendered images of the generated 3D human are most closely aligned with the ground truth. All reported scores demonstrate the superiority of the proposed Human-VDM over existing SOTA methods.

\subsection{User Study} 

The discussed metrics may not always fully capture the quality of generated 3D humans in terms of realism and other details. Thus following previous works, a user preference study was conducted to evaluate the performance of Human-VDM against existing SOTA methods. We compare Human-VDM with six recent SOTA models using 10 different samples, each with four views of generated 3D humans in different samples. For each sample, 30 volunteers were asked to vote on their impressions regarding four key aspects: geometry quality, texture quality, face quality, and overall quality. For a fair comparison, the results for the other six SOTA models were generated using their official code, with all settings left at their default values. As shown in Table~\ref{user_tab}, the proposed Human-VDM surpasses SOTA models in the aforementioned aspects. 

Most volunteers considered Human-VDM to generate the best results, especially in terms of geometry and texture. Though Human-VDM does not particularly dominate in face quality relatively, it performs the best face consistency with the input image as shown in Figure~\ref{comapre}. More than $53\%$ of the volunteers confirm that Human-VDM outperforms other SOTA models, which confirms Human-VDM's superiority.

\subsection{Ablation Study} 
\begin{figure}[!h]
\centering
\includegraphics[width=\columnwidth]{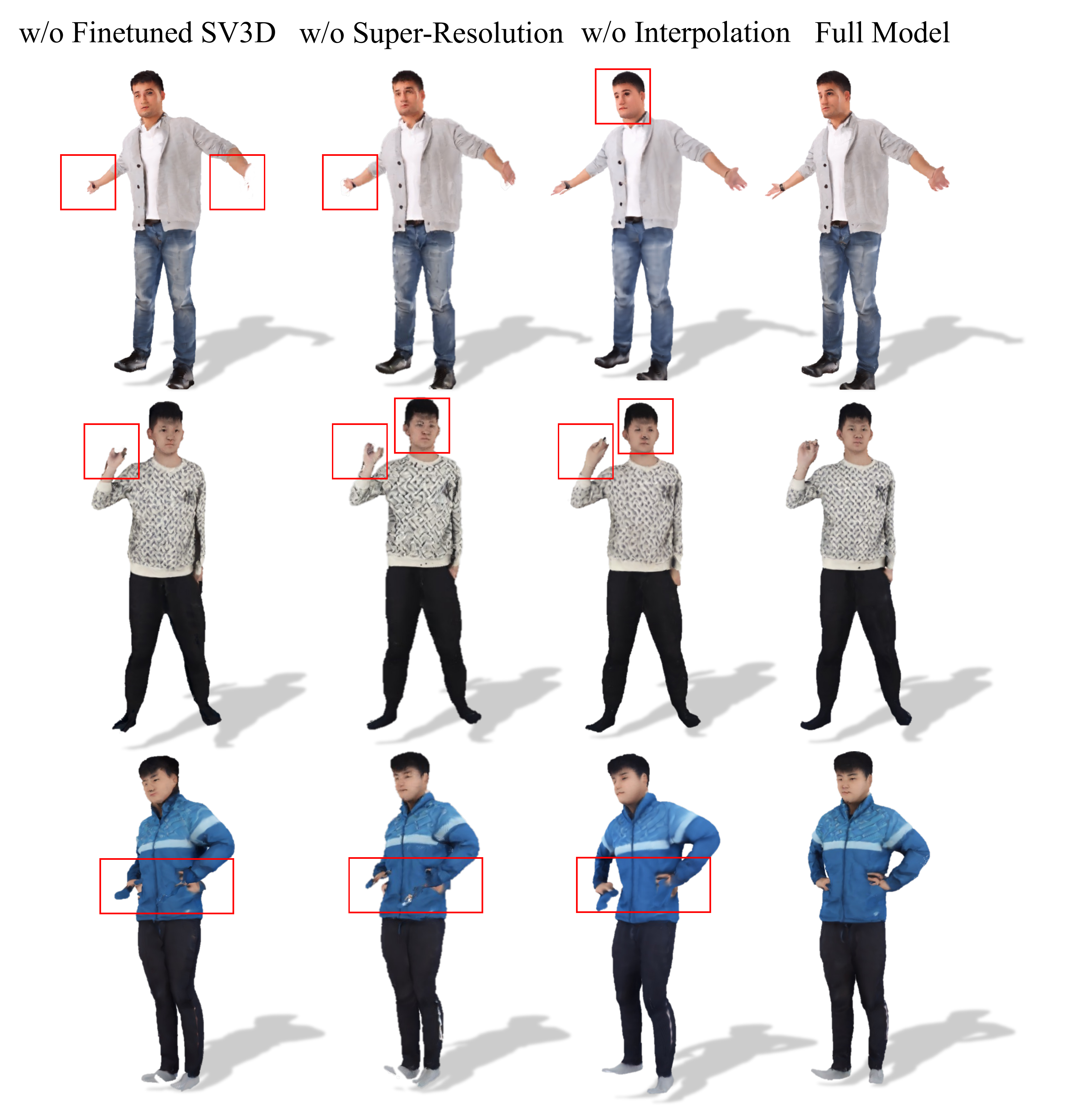}
\caption{\textbf{Qualitative Visual Ablation Comparisons.} Compared to other variants, the proposed full model achieves highly realistic textures and accurate geometry.}
\label{ablation}
\end{figure}

We performed ablation studies by systematically excluding various components to assess the effectiveness of the proposed modules through both quantitative and qualitative comparisons. For this analysis, we randomly selected $30$ samples from the Thuman~2.0 dataset~\cite{tao2021function4d}. We compared the full model with the variants excluding the proposed modules using the CLIP Similarity~\cite{radford2021learning}, SSIM~\cite{wang2004image}, LPIPS~\cite{zhang2018unreasonable}, and PSNR metrics. The evaluation covered rendered results from four viewpoints: front, back, left, and right. We additionally report results solely for the front view as well. Table~\ref{ablation_tab} presents the quantitative comparisons, while the qualitative visual comparisons are illustrated in Figure~\ref{ablation}.

Quantitative results demonstrate that the proposed full model achieves superior CLIP Similarity and SSIM across both the single view and four views. The visual ablation results further establish that the 3D human generated by the full model exhibits more photorealistic textures and precise geometry. Results produced without finetuned SV3D are less lifelike and realistic since the videos generated by the original SV3D are not satisfactory. Without Super-Resolution, the video frames are not distinct enough for the Human Gaussian Splatting module, which results in blurs and artifacts of the reconstructed humans. Due to the lack of features presented by only 21 frames, results generated without frame interpolation are not good enough yet, which has apparent artifacts in novel views. This confirms the significance and contribution of the video augmentation module. In general, the finetuned SV3D provides high-quality human orbital video for realistic reconstruction; the super-resolution module enhances the quality of video frames to generate more distinct results, and the VFI module enables the model to generate remarkable results in novel views. Although the full model shows a slight decrease in LPIPS and PSNR, the visual results indicate that the 3D human reconstructed by the complete model is of higher quality. Overall, the full model achieves better performance i.e., when including the proposed components. This confirms the effectiveness of the proposed modules.

\section{Conclusion and Future Work}

We propose a novel 3DGS-based framework for generating 3D humans from a single RGB image leveraging human video diffusion models. We first generate a view-consistent orbital video around the human and then augment the video through super-resolution and video frame interpolation. Finally, we reconstruct a remarkable 3D human using 3D Gaussian with the enhanced video. Both quantitative and qualitative experiments demonstrate that Human-VDM excels in generating 3D humans from a single image, outperforming state-of-the-art methods.
\vspace{0.5em}

\noindent \textbf{Limitations and Future works.} Human-VDM has two limitations. First, it is challenging to accurately generate precise finger geometry due to the intricate and small size of finger poses. Second, applying large video diffusion models limits the model's overall ability to achieve a real-time 3D human generation. Future works can focus on addressing these limitations by enhancing geometry generation for complex and small finger poses, as well as developing more efficient models that can achieve real-time 3D human generation.

\bibliography{aaai25}

\clearpage

\twocolumn[
\centering \textbf{\LARGE Supplementary Material}
\vspace{2em}
]

\noindent In the supplementary material, we provide a more detailed explanation of the model architecture, as well as training specifics, such as loss function weights, dataset descriptions, and definitions of the evaluation metrics. Additionally, we include further visual results and an analysis of failure cases.

\section{Model Architecture Details}

\subsection{Human Video Diffusion Module}

\noindent \textbf{Module Architecture.} The Video Diffusion Module of Human-VDM is based on  SV3D~\cite{voleti2024sv3d}. SV3D's architecture builds upon SVD~\cite{blattmann2023stable} and consists of a UNet~\cite{ronneberger2015u} model with multiple layers. Each layer comprises a sequence of 1 residual block with Conv3D layers, followed by spatial and temporal transformer blocks integrated with attention layers. After being embedded into the latent space via the visual autoencoder (VAE) of SVD, the conditioning image is concatenated with the noisy latent state input $z_t$ at noise timestep $t$ before being fed into the UNet. The CLIP-embedding~\cite{radford2021learning} matrix of the input image is provided to the cross-attention layers of each transformer block~\cite{vaswani2017attention}, serving as the key and value, with the layer's feature acting as the query. Along with the diffusion noise timestep, the camera trajectory is also incorporated into the residual blocks. The camera pose angles $e_i$ and $a_i$ are first embedded into the position embeddings. These camera pose embeddings are then concatenated, linearly transformed, and combined with the noise timestep embedding. The composite embedding is fed into every residual block, where it is added to the block’s output after another linear transformation to match the feature size.
\vspace{0.5em}

\noindent \textbf{Static Orbits.} The original SV3D model~\cite{voleti2024sv3d} consists of two main orbits: (1) the static orbit and (2) the dynamic orbit. Our study utilizes the static orbit, where the camera moves around the object at evenly spaced azimuth angles while maintaining the same elevation angle as in the conditioning image.
\vspace{0.5em}

\noindent \textbf{Fine-tuning SV3D for Human Video Diffusion.} The original SV3D is fine-tuned upon SVD-xt~\cite{blattmann2023stable} on the Objaverse dataset~\cite{deitke2023objaverse}, which contains synthetic 3D objects covering a wide diversity. For each object, ~\cite{voleti2024sv3d} renders $21$ frames around it on a random color background at $576\times576$ resolution, field-of-view of $33.8$ degrees. We adopt the same rendering strategy for the Thuman~2.0 dataset~\cite{tao2021function4d} to fine-tune SV3D for high-quality human video generation.

\subsection{Video Augmentation Module}

\textbf{Video Super-Resolution sub-module.} 
CodeFormer~\cite{zhou2022towards} is a transformer-based model~\cite{vaswani2017attention} to enhance the resolution of human images. Upon learning a discrete codebook, an encoder $E_H$ embed the high-quality human image $I_h \in \mathbb{R}^{H\times W\times 3}$ as a compressed feature $Z_h \in \mathbb{R}^{m\times n\times d}$ by an encoder $E_H$. Each ``pixel'' in $Z_h$ is then replaced by the nearest entry in the learnable codebook $\mathcal{C} = {c_k \in \mathbb{R}^d}^N_{k=0}$. Afterward, the quantized feature $Z_c \in \mathbb{R}^{m\times n\times d}$ along with the code token sequence $s \in {0, \cdots, N-1}^{m\cdot n}$ are produced as the following:
\begin{equation} \label{sf1}
\begin{split}
    Z_c^{(i,j)}&=\arg\min\limits_{c_k\in\mathcal{C}}\|Z_h^{(i,j)}-c_k\|_2,\\
    \quad s^{(i,j)}&=\arg\min\limits_k\|Z_h^{(i,j)}-c_k\|_2.
\end{split}
\end{equation}
Given $Z_c$, the high-quality human image $I_{rec}$ is reconstructed by the decoder $D_H$. The $m \times n$ code token sequence, denoted as $s$, constitutes a novel latent discrete representation, which encodes the specific indices corresponding to entries in the learned codebook, i.e., $Z^{(i,j)}_c = c_k$ when $s^{(i,j)} = k$.

Subsequently, with the codebook $\mathcal{R}$ and decoder $D_H$ held constant, a Transformer module~\cite{vaswani2017attention} is introduced for predicting the code sequence, capturing the global human composition from low-quality inputs. To extract the low-quality features $Z_l \in \mathbb{R}^{m \times n \times d}$ using $E_L$, the features are first unfolded to $m \cdot n$ vectors $Z_l^v \in \mathbb{R}^{(m \cdot n) \times d}$, which are subsequently fed into the Transformer. In the transformer, the $s^{th}$ self-attention block performs the below operation:
\begin{equation} \label{sf2}
\begin{split}
    X_{s+1}=\text{Softmax}(Q_sK_s)V_s+X_s,
\end{split}
\end{equation}
where $X_0 = Z^v_l$. $X_s$ is used to get the queries $Q$, key $K$, and value $V$ through linear layers.
\vspace{0.5em}

\noindent\textbf{Video Frame Interpolation (VFI) sub-module.} PerVFI is a novel model of frame interpolation. Given two reference frame images, $I_0$ and $I_1 \in \mathbb{R}^{H\times W\times 3}$, with height $H$ and width $W$, PerVFI is designed for reconstructing the intermediate frame $I_t$ within the target time $t \in (0, 1)$. It incorporates an asymmetric synergistic blending (ASB) module and a conditional normalizing flow-based generator.

After estimating bidirectional optical flows, PerVFI presents a pyramidal architecture, which can better capture multiscale information to extract features at different scales.
Specifically, a feature encoder $E_{\theta}$ is used to encode the two images into pyramid features with $L$ levels, which can be denoted as $f_i=E_{\theta(I_i)}$, $i=0,1$. Subsequently, a feature blending module, denoted as $B_\theta$, blends the pyramidal features to produce intermediate pyramid features. Afterward, a conditional normalizing flow-based generator $G_\phi$, which is invertible, decodes $f_t$ into the output frame $I_t$. The output is formulated as $I_t=G^{-1}_\phi(r;f_t)$, where $r \sim \mathcal{N}(0,\tau) \in \mathbb{R}^{H \times W \times 3}$ represents a variable drawn from a normal distribution with a temperature parameter $\tau$; $f_t$ is the feature pyramid with $L$ levels.

\begin{figure*}[!h]
\centering
\includegraphics[width=2\columnwidth]{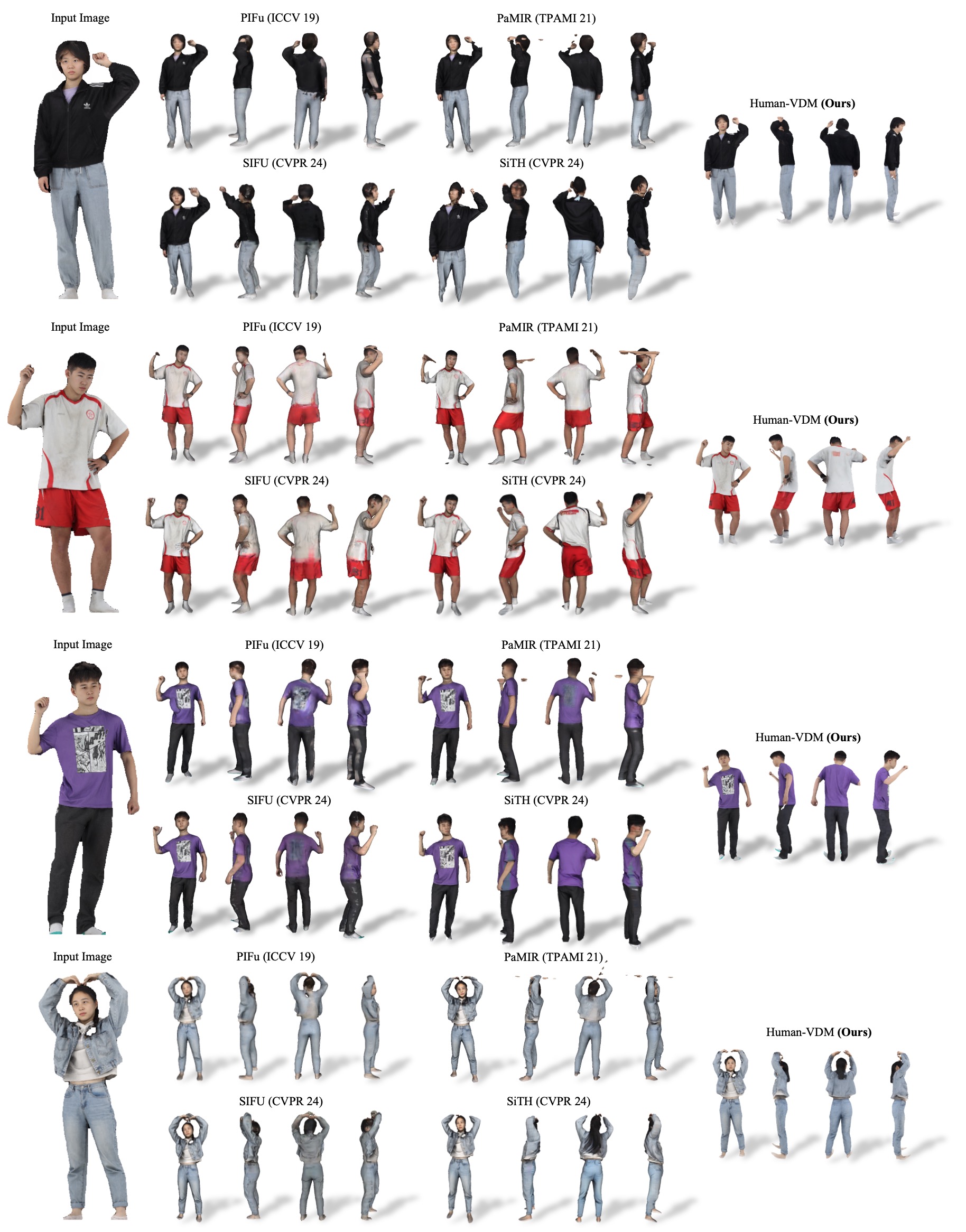}
\caption{Additional results comparing Human-VDM with SOTA models. The results demonstrate that Human-VDM achieves superior 3D human generation quality. \faSearch~\textbf{zoom in} for details.}
\label{supp1}
\end{figure*}

\begin{figure*}[!h]
\centering
\includegraphics[width=1.95\columnwidth]{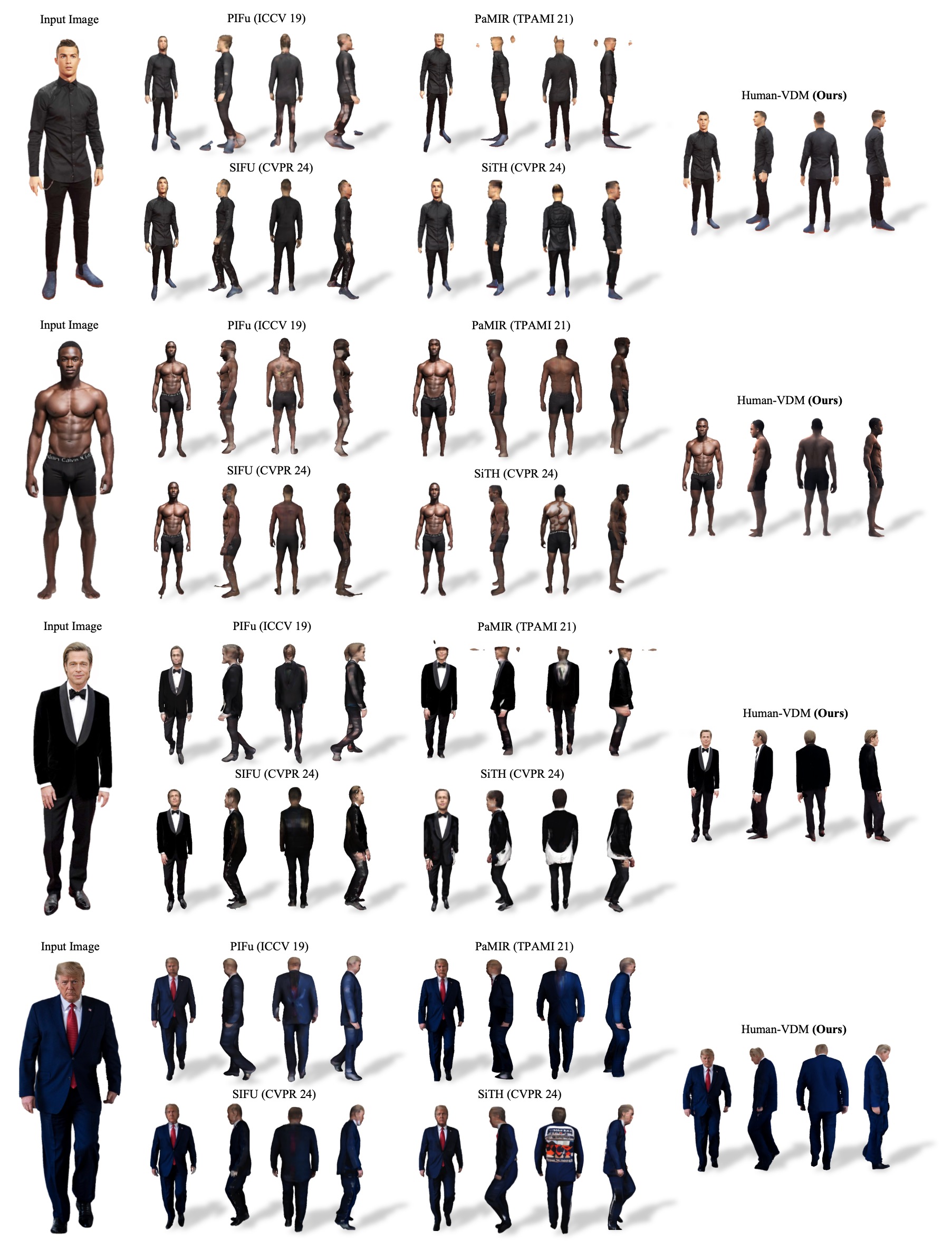}
\caption{In-the-wild testing results comparing Human-VDM with SOTA models. The results demonstrate that Human-VDM achieves superior 3D human generation quality. \faSearch~\textbf{zoom in} for details.}
\label{supp2}
\end{figure*}

\begin{figure*}[!h]
\centering
\includegraphics[width=2\columnwidth]{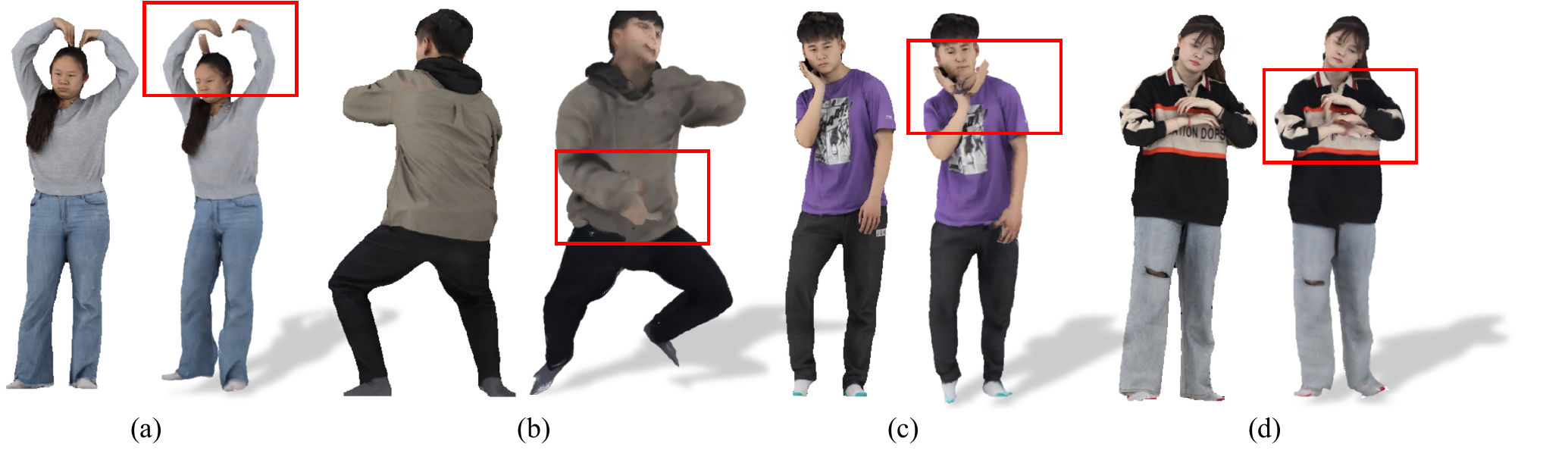}
\vspace{-1em}
\caption{Failure cases of Human-VDM. The intricate and small size of fingers makes it challenging to accurately generate precise finger geometry, as shown in (a) and (b). Moreover, we can see the case of (c) hand-face and (d) hand-hand interactions, which remain challenging in 3D human generation. The left image shows the input, while the right is the generated 3D human.}
\label{supp3}
\end{figure*}

\subsection{3D Human Gaussian Splatting Module}

In 3D Gaussian, human appearances are determined by point displacements $dT$ and properties $\textbf{P}$. Modeling dynamic human appearances involves estimating these evolving properties. We propose a dynamic appearance network coupled with an optimizable feature tensor to effectively capture dynamic human appearances across various poses. The dynamic appearance network is designed to learn a mapping from a 2D manifold representing the underlying human shape to the dynamic properties of 3D Gaussians as follows:
\begin{equation} \label{sf4}
\begin{split}
    f_\phi:\mathcal{S}^2\in\mathbb{R}^3\to\mathbb{R}^7,
\end{split}
\end{equation}
the 2D human manifold $\mathcal{S}^2$ is depicted by a UV positional map $I \in \mathbb{R}^{H\times W \times3}$, where each valid pixel stores the position $(x, y, z)$ of one point on the posed body surface. The final predictions consist of per point offset $\Delta\hat{\mathbf{x}}\in\mathbb{R}^3,\text{ color }\hat{\mathbf{c}}\in\mathbb{R}^3,\text{ and scale }\hat{s}\in\mathbb{R}$.

Human poses $\boldsymbol{\theta}$ and translations $t$ estimated from monocular videos are usually inaccurate. Hence, the 3D Gaussians reposed in motion space may be inaccurately represented, potentially resulting in unsatisfactory rendering outcomes. To address this issue, we jointly optimize human motions and appearances. We update the estimated body poses and translations by calculating $(\Delta\boldsymbol{\theta},\Delta\mathbf{t})$ to refine human motions, which can be formulated as follows:
\begin{equation} \label{sf5}
\begin{split}
    \hat{\boldsymbol{\Theta}}=(\boldsymbol{\theta}+\Delta\boldsymbol{\theta},\mathbf{t}+\Delta\mathbf{t}).
\end{split}
\end{equation}
We modify $\theta$ in the equation of animatable Gaussians in the main article using $\hat{\boldsymbol{\Theta}}$ to render the proposed animatable 3D Gaussians differentiable with respect to the motion conditions. Finally, the current frame image is taken as the ground truth to calculate the loss with the rendered image. 

\subsection{Training Objectives}
We use the current frame image, i.e., $f_i$, and the rendered image, i.e., $f_i^r$, for supervising the Human-VDM model. The total loss consists of six different loss functions which include $\mathcal{L}_{\text{RGB}}$, $\mathcal{L}_{\text{SSIM}}$, $\mathcal{L}_{\text{LPIPS}}$, $\mathcal{L}_{\text{Offset}}$, $\mathcal{L}_{\text{Scale}}$ and $\mathcal{L}_{f}$. In this section, we describe the loss functions in greater detail.\\

\noindent$\mathcal{L}_{\text{RGB}}$ is the L1-loss between the ground truth and the rendered frame and is formulated as:
\begin{equation} \label{sf6}
\begin{split}
    \mathcal{L}_{\text{RGB}}(x,y)=\frac{1}{HW}\sum_{h,w}^{HW}|y_{hw}-x_{hw}|,
\end{split}
\end{equation}

\noindent$\mathcal{L}_{\text{SSIM}}$~\cite{wang2004image}, or the Structural Similarity Index Metric Loss is a perceptual metric to measure the similarity between two images, taking luminance, contrast, and structure into account. We define the SSIM loss as follows:

\begin{equation} \label{sf7}
\begin{split}
    \mathcal{L}_{\text{SSIM}}(x,y) &= 1 - \text{SSIM}(x,y) \\
    &=1 - \frac{(2\mu_{x}\mu_{y}+c_{1})(2\sigma_{xy}+c_{2})}{(\mu_{x}^{2}+\mu_{y}^{2}+c_{1})(\sigma_{x}^{2}+\sigma_{y}^{2}+c_{2})},
\end{split}
\end{equation}
where $\mu_x \text{and } \mu_y$ stands for the mean of $x$ and $y$; $\sigma_x$ and $\sigma_y$ represent the variance of $x$ and $y$, while $\sigma_{xy}$ denote the covariance of $x$ and $y$.

\noindent$\mathcal{L}_{\text{LPIPS}}$~\cite{zhang2018unreasonable} measures image similarity, which evaluates the perceptual difference between two images through deep learning models. In this paper, we utilize AlexNet~\cite{krizhevsky2012imagenet} for extracting features of images. We calculate $\mathcal{L}_{\text{LPIPS}}$ as:
\begin{equation} \label{sf8}
\begin{split}
  \mathcal{L}_{\text{LPIPS}}(x,y)=\sum_l\frac1{H_lW_l}\sum_{h,w}||w_l\odot(\hat{f}_{xhw}^l-\hat{f}_{yhw}^l)||_2^2,
\end{split}
\end{equation}
where $\hat{f}_{xhw}^l$ represents the feature output of image $x$ in layer $l$ at the pixel $hw$, and $\hat{f}_{yhw}^l$ means the same of image $y$. $w_l$ is a trainable parameter in layer $l$.

\noindent$\mathcal{L}_{\text{Offset}}$, $\mathcal{L}_{\text{Scale}}$ and $\mathcal{L}_{\text{f}}$  calculate the L2-norm of the feature map, predicted offsets and scales on the canonical surface, respectively. We formulate them as follows:
\begin{equation} \label{sf9}
\begin{split}
  \mathcal{L}_{\text{Offset}}=\frac1N\sum_{i=1}^N(\Delta \hat{x_i})^2,
\end{split}
\end{equation}
where $\Delta x_i$ denote the predicted offset of $i^{th}$ gaussian.
\begin{equation} \label{sf10}
\begin{split}
  \mathcal{L}_{\text{Scale}}=\frac1N\sum_{i=1}^N(\hat{s_i})^2,
\end{split}
\end{equation}
where $s_i$ denotes the predicted scale of $i^{th}$ gaussian.
\begin{equation} \label{sf11}
\begin{split}
  \mathcal{L}_{\text{f}}=\frac1F\sum_{i=1}^F(t_i)^2,
\end{split}
\end{equation}
where $t_i$ denotes the optimized feature.

\section{Implementation Details}

In this section, we present additional details on the model implementation. The Gaussian decoder is implemented as an MLP. A total of 202,738 Gaussians were initially sampled on the surface of the canonical SMPL model. The adjustable coefficient $w$, which presents the reliance on input low-quality image, is set to $0.7$ in the Super-Resolution module. For each sample, we train the dynamic appearance network on a single NVIDIA RTX 3090 GPU for 1000 epochs with a batch size of 2. The learning rate of the network is set to \texttt{3e-3}.

\section{Additional Results}

In this section, we present additional results, including in-the-wild testing and failure cases.

\subsection{In-the-wild visual results}

To demonstrate the superiority of Human-VDM, we provide more visual comparison results. This includes additional results as shown in Figure~\ref{supp1}, including results on challenging in-the-wild cases illustrated in Figure~\ref{supp2}.

\subsection{Failure Cases}
In this subsection, we present several cases of failure in Human-VDM.
Although Human-VDM performs exceptionally well in generating 3D humans from a single RGB image, it still has a few limitations and failure cases, as discussed in the main text. Figure~\ref{supp3} shows the failure cases of Human-VDM. For example, when the human in the input image interacts with their hands against their body, some artifacts may appear at the contact region.

\end{document}